\documentclass{article} 
\usepackage{iclr2027_conference,times}

\usepackage{amsmath,amsfonts,bm}

\def\eqref#1{equation~\ref{#1}}

\def\1{\bm{1}}

\DeclareMathAlphabet{\mathsfit}{\encodingdefault}{\sfdefault}{m}{sl}
\SetMathAlphabet{\mathsfit}{bold}{\encodingdefault}{\sfdefault}{bx}{n}

\usepackage{graphicx}
\usepackage{hyperref}
\usepackage{booktabs} 
\usepackage{url}
\usepackage{multirow} 
\usepackage{caption}

\title{ScenePlacer: Reformulating Scene-Level Object Pose Estimation as Visibility Field Learning}

\author{\textbf{Ken Deng}$^{1,2}$ \quad
\textbf{Jiarui Liu}$^{1,2}$ \quad
\textbf{Guibiao Liao}$^{2,3}$ \quad
\textbf{Kunming Luo}$^{1,2}$ \quad
\textbf{Mochu Xiang}$^{2,3}$ \\
\textbf{Hongyu Yan}$^{1,2}$ \quad
\textbf{Yebin Liu}$^{4}$ \quad
\textbf{Shenghua Gao}$^{2,3}$ \quad
\textbf{Yizhou Yu}$^{2,3}$ \quad
\textbf{Ping Tan}$^{1,2}$ \\
$^{1}$The Hong Kong University of Science and Technology
\quad
$^{2}$Shenzhen Loop Area Institute\\
$^{3}$The University of Hong Kong
\quad
$^{4}$Tsinghua University
}

\usepackage{xcolor}
\definecolor{commentgreen}{RGB}{20,140,70}
\definecolor{commentpurple}{RGB}{130,60,180}

\usepackage[normalem]{ulem}

\iclrfinalcopy 

\begin{document}

\maketitle

\lhead{Preprint}

\begin{center}
    \vspace{-3em}
    \includegraphics[width=\textwidth]{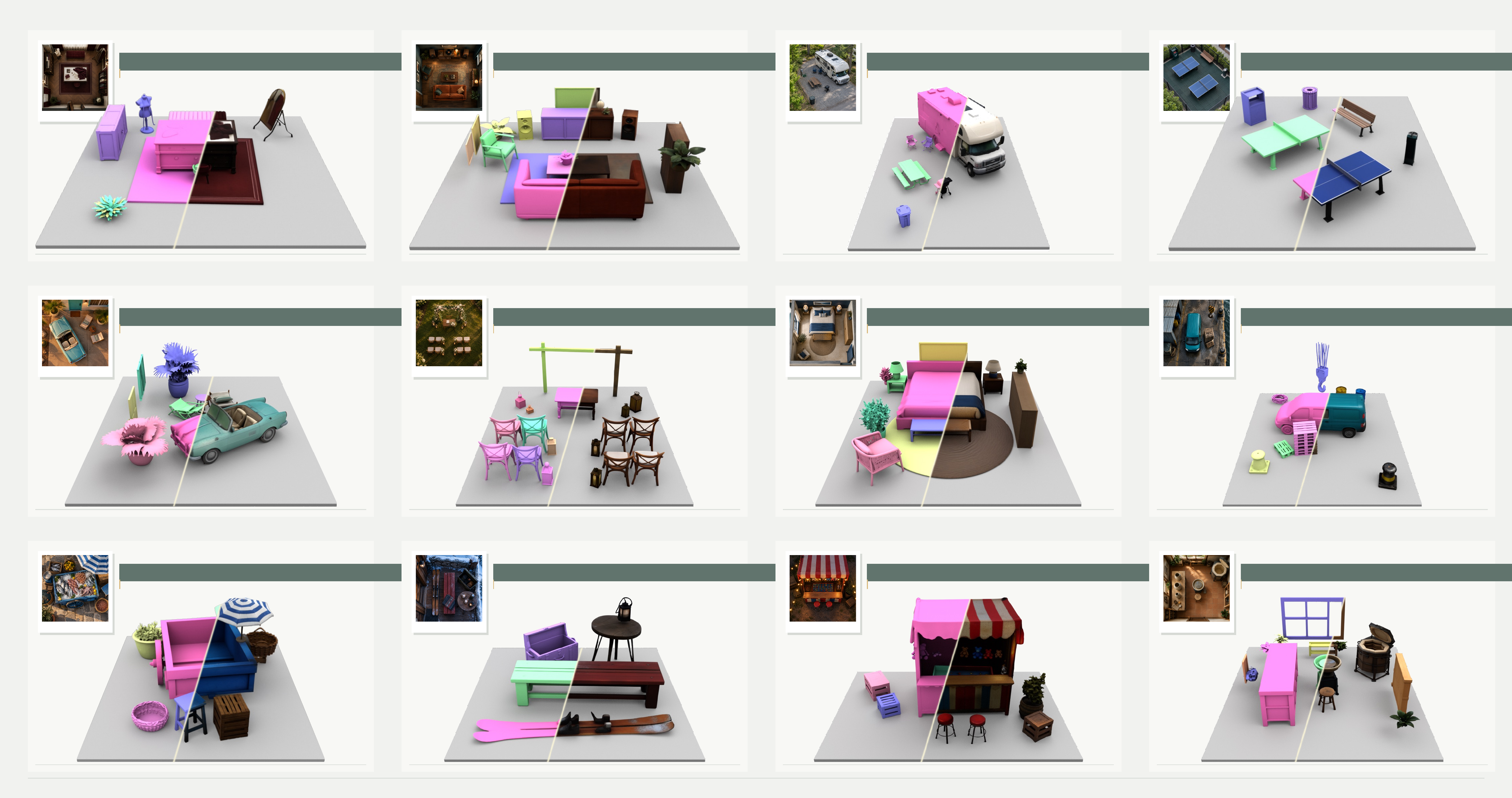}
    \captionof{figure}{Starting from an input RGB image, and given the generated object and its estimated geometry (different colors represent for different parts), our method robustly and efficiently estimates the object's pose within the scene.}
    \label{fig:teaser}
\end{center}

\begin{abstract}
We present ScenePlacer, a framework for scene-level pose estimation of objects generated by state-of-the-art native shape generation models, placing generated 3D objects into scenes given a partial geometric observation without modifying the underlying generative pipeline. Existing scene generation approaches either generate objects directly in scene coordinates, compromising their canonical-space priors, or regress poses from complete generated shapes in the shape generation process — the latter being a poorly conditioned problem that is sensitive to scale ambiguity and partial observations. ScenePlacer decouples object generation from pose estimation by reformulating alignment as conditional visibility-field learning. Given an observed partial point cloud from reconstruction and a generated mesh, it predicts a visibility field defined in object space, then estimates the rigid transformation by aligning the partial observation to visible parts on the mesh. We propose a pretrained 3D shape VAE as our backbone, preserving its mesh prior while conditioning it on a separately encoded observation. We further introduce visibility-aware, radius-constrained supervision and synthesize training pairs using only object-level meshes and rendered views, eliminating the need for scene-level pose annotations. On 3D-FUTURE, ScenePlacer consistently outperforms existing placement and pose-refinement baselines under both ground-truth and monocular depth estimation. With generated objects, it achieves a Chamfer distance of 0.0077 and an F-score of 92.40 under estimated depth and improves to 0.0023 and 98.27, respectively, under ground-truth geometry.
\end{abstract}
    
\section{Introduction}
\label{sec:intro}


Compositional 3D scene generation is essential for AR/VR and embodied AI. Recent 3D object generative models provide strong priors for recovering complete geometry from sparse observations, which can be transferred to this task. Thanks to the capabilities of 3D generative models~\citep{3dshape2vecset, trellis}, scene quality has improved significantly compared to before, yet there is still a considerable gap before the results are truly satisfactory, coming with degraded shape quality, excessive memory consumption, and inaccurate pose estimation.

Several methods generate objects within the scene in place. One strategy~\citep{midi} generates all objects in a shared canonical space, inevitably degraded due to limited resolution. The other strategy~\citep{shaper, 3dfixer} tries to inject the observed partial point cloud as a condition into the 3D generative model and generates each object sequentially. However, the observed point clouds are incomplete, and their orientation varies with the camera viewpoint. By requiring the model to generate objects with arbitrary translations and rotations, this approach violates the common assumption in 3D generative models that objects are centered at the origin and aligned with the coordinate axes, inevitably degrading generation quality.
An alternative approach~\citep{scenegen, sam3d} preserves the canonical generation space and introduces a separate pose prediction head that estimates the transformation (translation, rotation, and scale) from object to the world coordinate frame. Although this maintains high-quality object generation, pose prediction accuracy remains a major concern: Small perturbations in the transformation matrix can lead to large geometric deviations once applied, rendering the regression problem ill-conditioned.




Instead of struggling balancing the generation quality and pose estimation accuracy like existing methods, we view placing generated objects back into the scene as a distinct problem that warrants dedicated investigation. This problem presents challenges beyond conventional pose estimation. A generated object may preserve the structural or functional characteristics of the observed object without faithfully reproducing its visible geometry, making rigid registration unreliable. Moreover, generated objects are typically normalized to $[-1,1]^3$ and lack scene-scale information, introducing scale ambiguity that further complicates rendering-based pose optimization. Any6D~\citep{any6d} addresses this problem through multi-stage pose and scale refinement, but relies on close geometric agreement between the generated object and the observation, with errors potentially propagating and accumulating across refinement stages.
We propose a dedicated single-pass pose alignment model that operates on the generated mesh, enabling efficient and robust pose estimation without affecting generation quality. Specifically, given an observed partial point cloud $\mathcal{P}_{\text{obs}}$ of an object from reconstruction and its corresponding generated mesh $\mathcal{M}_{\text{gen}}$, we propose to predict the mesh-aligned visible surface $\mathcal{P}_{\text{aligned}}$ and solve for the transformation between $\mathcal{P}_{\text{obs}}$ and $\mathcal{P}_{\text{aligned}}$. 

Our method is based on the insight that pose estimation can be solved by finding the visible surface on a mesh in object canonical space corresponding to the observed part. This reframing removes the dependency on scene-level data, which remains scarce in practice.  Architecturally, we observe that our task---predicting a partial surface aligned with a given complete mesh---naturally aligns with the prior embedded in 3D Shape VAE, which encodes and decodes complete shapes. By treating the observed partial point cloud as a conditioning signal that modulates the mesh decoder, our network only needs to learn to select the visible subset of the mesh surface. This design enables strong generalization with only lightweight tuning. Our approach is designed to be general-purpose and independent of the underlying generation pipeline. It does not require modifications to the generated object or retraining of the generator, enabling straightforward integration with diverse object generators and geometry models.

In summary, our contributions are as follows:
\begin{itemize}
    \item We recast object estimation as learning a \emph{visibility field} with respect to the complete object mesh, transforming the pose estimation problem into a conditional surface prediction task that is inherently object-centric and independent of scene-level context.
    \item We introduce a lightweight pose estimation module that can be seamlessly integrated with existing generative models. It predicts the mesh-aligned visible surface and resolves the rigid transformation between the observed partial point cloud and the generated object, without altering the generation process.
    \item We propose a self-supervised training paradigm that relies solely on object-level data, repurposing a 3D shape VAE to address object placement in compositional scene generation, thereby eliminating the need for scarce scene-level supervision.
\end{itemize}
\section{Related Work}
\label{sec:related_work}

\paragraph{Geometry Foundation Models.} 
Recent geometry foundation models adopt transformer architecture for traditional 3D vision tasks such as depth estimation, camera pose estimation, and 3D point tracking. Some of them focus on single-view depth estimation~\citep{da, da2, djy_1, moge, moge2, ppd, marigold}, aiming to learn RGB-to-Depth transformation, while they can't handle multi-view input. Dust3R~\citep{dust3r} tried to handle stereo reconstruction by formulating the problem as direct regression of a point map for the second view in the first view's camera coordinate system. Moving forward, VGGT and a series of subsequent works~\citep{vggt, sailrecon, pi3, da3, glob3r} tried to extend to multi-view input: VGGT directly predicts per-view depth maps and per-view camera poses; DA3 adopts ray map as camera representation; Pi3 focuses on learning pose-invariant 3D representations that are agnostic to camera motion; Sail-Recon extends VGGT to long sequence input. However, the above works can only handle static scenes. Recent geometry foundation model D4RT~\citep{d4rt} focuses on 4D scene reconstruction.
\vspace{-3mm}
\paragraph{Scene Generation}

Scene generation approaches fall into two categories. The first leverages 3D object generation priors~\citep{hunyuan3d, clay, craftman, triposg, trellis, detailgen3d, hi3dgen, CTR3D, posemaster, sparc3d, direct3ds2, trellis2}, together with geometry foundation and pattern recognition priors~\citep{geosam2, sam3, groundedsam}, and extends them to decomposable scene generation~\citep{gen3dsr, midi, sam3d, scenemaker, 3dfixer, cast, scenegen, shaper}; in such scenes, objects are interactive and compatible with physics simulation. Some~\citep{layoutgpt, holodeck, holodeck2, mesatask, tabletopgen, scenesmith} use language models to orchestrate scenes from text prompts, producing more orderly layouts but lacking realism and image alignment. The second utilizes 2D image or video priors and incorporates 3D conditions (e.g., camera parameters) to generate multi-view image or video sequences with 3D consistency~\citep{gen3r, ctrlroom, spatialcrafter, wonderland, wonderworld, bolt3d, flashworld, cat3d, stereoworldmodel, worldr1,lyra2}. While these methods achieve superior visual quality via 2D priors, lacking native 3D representation makes ensuring 3D interaction and physical consistency difficult.
\vspace{-3mm}
\paragraph{Point Cloud Registration}
Point cloud registration is closely related to single-reference unseen object pose estimation, since both tasks depend on establishing reliable correspondences between partial 3D observations. Classical approaches typically operate on hand-crafted geometric cues. For instance, ICP and its variants~\citep{besl1992method, rusinkiewicz2001efficient} alternate between nearest-neighbor association and rigid transform refinement, while FPFH~\citep{rusu2009fast} characterizes local surface geometry via point feature histograms, and PPF~\citep{drost2010model} represents oriented point-pair relations for generating pose hypotheses. Although these methods serve as reasonable geometric baselines, they are often fragile under outliers, repetitive structures, and limited view overlap. Learning-based techniques attempt to overcome these limitations by predicting more distinctive correspondence signals. Predator~\citep{huang2021predator}, for example, introduces overlap-aware feature learning so that matching concentrates on regions that are actually shared between two partially overlapping scans. Similarly, GeoTransformer~\citep{Geotransformer} aggregates pairwise distances and angular information into rotation-invariant geometric features, thereby enabling robust superpoint matching across large viewpoint changes. More recently, Any6D~\citep{any6d} addresses this setting in a model-free manner by reconstructing a normalized object shape from a single RGB-D anchor image, jointly aligning it to metric scale in 2D/3D, and selecting render-and-compare pose hypotheses for the query image.
\section{Preliminary: 3DShape2VecSet Representation}
We build on the 3DShape2VecSet~\cite{3dshape2vecset} representation, which encodes a 3D shape as a set of latent vectors $\{\mathbf{f}_i\}_{i=1}^M$ without explicit spatial coordinates. Given an input point cloud with coordinates $\mathbf{X}$ and normals $\mathbf{N}$, the encoder uses cross-attention to aggregate per-point features into $M$ latents:

\begin{equation}
\{\mathbf{f}_i\} = \mathrm{CrossAttn}\bigl(\mathrm{PosEmb}(\mathbf{X}_0, \mathbf{N}_0),\; \mathrm{PosEmb}(\mathbf{X}, \mathbf{N})\bigr),
\end{equation}
where $\mathbf{X}_0$ is obtained by farthest point sampling from $\mathbf{X}$, and $\mathbf{N}_0$ denotes the normals at the sampled points. For decoding, an occupancy value at query coordinate $\mathbf{x}$ is predicted by attending from $\mathbf{x}$ to the latent set:
\begin{equation}
\hat{\mathcal{O}}(\mathbf{x}) = \mathrm{Linear}\!\left(\sum_i \mathbf{v}(\mathbf{f}_i) \cdot \mathrm{softmax}\!\left(\mathbf{q}(\mathbf{x})^\top \mathbf{k}(\mathbf{f}_i)\right)\right).
\end{equation}
This defines a continuous occupancy field that can be queried at arbitrary points.
\section{Method}
\label{sec:method}

\begin{figure*}[t]
    \centering
    \includegraphics[page=1, width=\textwidth]{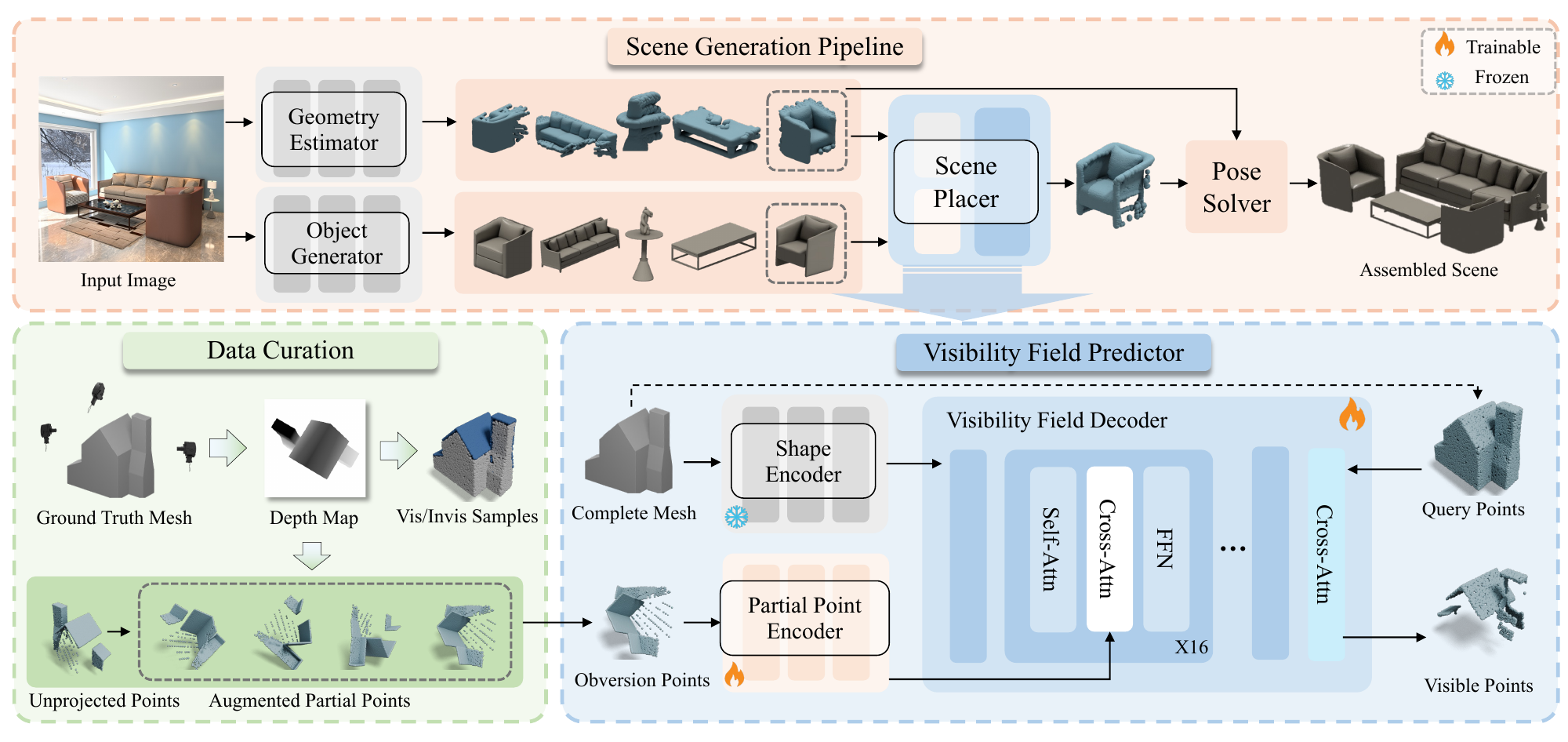}
    \caption{\textbf{Overview.} \textit{Pipeline:} The input image is first processed by a geometry estimator to obtain a point cloud and a mesh generated by a generative model; ScenePlacer is then used to place the object back into its corresponding spatial location, resulting in a compositional scene. \textit{Data Curation:} Depth maps rendered from the ground-truth mesh are back-projected into oriented partial points, augmented by random translation, rotation and scales, and vis/invis samples with an exclusion radius provide supervision. \textit{Visibility Field Predictor:} A frozen Shape Encoder and a trainable Partial Point Encoder produce mesh and observation latents, respectively; the Visibility Field Decoder conditions the mesh latents on the observation, and decodes visibility for query points sampled on the mesh surface, yielding the visible points used for alignment.}
    \label{fig:pipeline}
\end{figure*}

Let $\mathcal{M}$ denote the object mesh and $\mathcal{P}_{\text{obs}}$ an observed partial point cloud reconstructed from image(s) or scanning. The observed cloud is initially unaligned with the mesh and contains only 3D coordinates. We reformulate the alignment task as \emph{visibility field learning}: a network $F$ takes the mesh and the observation as input and predicts, for any query point $\mathbf{x} \in \mathbb{R}^3$, the probability that $\mathbf{x}$ lies on the visible surface of $\mathcal{M}$ as seen from the observation viewpoint. Formally,
\begin{equation}
F(\mathcal{M}, \mathcal{P}_{\text{obs}}): \mathbb{R}^3 \to [0,1], \quad \hat{\mathcal{V}}(\mathbf{x}) = F(\mathcal{M}, \mathcal{P}_{\text{obs}})(\mathbf{x}),
\end{equation}
where $\hat{\mathcal{V}}(\mathbf{x})$ denotes the predicted visibility. The aligned visible surface $\mathcal{P}_{\text{aligned}}$ is obtained by sampling points with $\hat{\mathcal{V}}(\mathbf{x}) > \tau$. The rigid transformation between the mesh and the observed cloud is then recovered by aligning the two partial point sets:
\begin{equation}
\mathcal{T}^* = \arg\min_{\mathcal{T}} \operatorname{CD}\bigl(\mathcal{T}(\mathcal{P}_{\text{aligned}}),\; \mathcal{P}_{\text{obs}}\bigr),
\end{equation}
where $\mathcal{T}$ denotes a rigid transformation and $\operatorname{CD}$ denotes the chamfer distance between two point sets. The network is trained with binary cross-entropy supervision: positive queries are sampled from the rendered visible surface, while negative queries are drawn from the mesh surface with an exclusion radius that ensures they correspond to genuinely occluded or unobserved regions (Sec.~\ref{subsec:supervision}). Throughout this pipeline, point clouds serve as the sole geometric representation. The overview of our method can be found in Figure~\ref{fig:pipeline}.

\subsection{Observation Latent Encoding}
\label{subsec:obs_encoding}

To encode the observed point cloud and the complete point cloud into a shared latent space, we also represent the encoding of the observed point cloud in the form of a Vecset and initialize both encoders using the same weights. We encode the partial observation and the complete mesh with separate encoders: the mesh encoder is frozen from a pretrained encoder, while the observation encoder adopts the same architecture but is fully trainable, preventing the partial observation from disturbing the mesh prior. The observed point cloud $\mathcal{P}_{\text{obs}}$ contains only 3D coordinates and no normals. We first downsample it to a subset $\mathcal{P}_0$ via farthest point sampling, then use cross-attention to aggregate positional embeddings from both sets into $M$ observation latents:
\vspace{-1mm}
\begin{equation}
\{\mathbf{o}_i\} = \mathrm{CrossAttn}\bigl(\mathrm{PosEmb}(\mathcal{P}_0),\; \mathrm{PosEmb}(\mathcal{P}_{\text{obs}})\bigr).
\end{equation}
In contrast to the original 3DShape2VecSet formulation, we omit the KL regularization bottleneck. Since our goal is conditional reconstruction rather than generative modeling, retaining the full latent capacity helps preserve fine geometric detail and reduces the risk of entangling observation noise with structural information in a compressed latent space.

\subsection{Visibility Field Predictor}
\label{subsec:visibility_decoding}

The mesh decoder is originally trained to reconstruct complete, watertight surfaces from mesh latents alone, and thus provides a strong geometric prior. We do not discard this prior but instead use the observation latents to condition it. Specifically, we interleave cross-attention layers into the decoder's self-attention stack. Let $\mathbf{M} = \{\mathbf{m}_i\}_{i=1}^{M}$ denote the mesh latents and $\mathbf{O} = \{\mathbf{o}_i\}_{i=1}^{M}$ the observation latents. Each decoder block is defined as:
\begin{equation}
\mathbf{M}^{(l+1)} = \mathrm{CrossAttn}\bigl(\mathrm{SelfAttn}(\mathbf{M}^{(l)}),\; \mathbf{O}\bigr),
\end{equation}
where the mesh latents act as queries and the observation latents provide keys and values. This asymmetric design allows the mesh latents to retain their original representational capacity while being modulated by observed evidence. The decoder therefore does not regress the visible surface from scratch; instead, it biases its existing mesh reconstruction toward the region that is consistent with the observation. To ensure stable initialization, the final projection is zero-initialized, so that at the start of training the conditioning path introduces no distortion and the model initially behaves as the unconditional mesh decoder.

After obtaining the modulated mesh latents $\mathbf{M}^{(L)}$, we decode visibility by sampling query points directly on the mesh surface. For each query point $\mathbf{x} \in \mathcal{S}_{\mathcal{M}}$, we compute
{\small
\begin{equation}
\hat{v}(\mathbf{x}) = \sigma\Bigl(\mathrm{Linear}\Bigl(\sum_i \mathbf{v}(\mathbf{m}_i^{(L)}) \cdot \mathrm{softmax}\!\left(\mathbf{q}(\mathbf{x})^\top \mathbf{k}(\mathbf{m}_i^{(L)})\right)\Bigr)\Bigr),
\end{equation}
}
where $\sigma$ is the sigmoid function and $\mathbf{q},\mathbf{k},\mathbf{v}$ are learnable projections. The predicted visible surface is obtained by thresholding:
\begin{equation}
\mathcal{P}_{\text{aligned}} = \{ \mathbf{x} \in \mathcal{S}_{\mathcal{M}} \mid \hat{v}(\mathbf{x}) > 0.5 \}.
\end{equation}

\subsection{Visibility-Aware Supervision}
\label{subsec:supervision}

The visibility field is supervised in a self-supervised manner. Visible queries are sampled directly from the canonical visible surface $\mathcal{P}_{\text{aligned}}$ and assigned label $y=1$. Invisible queries are sampled from the full mesh surface, but are accepted only if they lie outside an exclusion radius $r_{\text{excl}}$ from any visible point. Formally, let $\mathcal{R}$ be a reference subset of $\mathcal{P}_{\text{aligned}}$. A candidate invisible point $p_{\text{inv}}$ is accepted only if
\begin{equation}
\min_{r \in \mathcal{R}} \| p_{\text{inv}} - r \|_2 > r_{\text{excl}}.
\end{equation}
This constraint is critical for learning a correct visible/invisible separation. If invisible queries were sampled uniformly from the mesh surface, many would fall near the boundary of the visible region or in the gaps between visible surface patches. These points are geometrically close to visible points but semantically belong to the occluded or unobserved part of the object. Without the exclusion radius, the network receives contradictory supervision and tends to interpolate between visible and invisible regions instead of learning a sharp visibility boundary. By enforcing a hard radius constraint around visible points, we ensure that invisible queries correspond unambiguously to non-visible surface regions, which stabilizes training and improves the quality of the predicted visible surface.

 
Let $\mathcal{Q} = \mathcal{Q}^+ \cup \mathcal{Q}^-$ denote the set of all accepted query points. The network is trained with a binary cross-entropy loss:

\begin{equation}
\begin{aligned}
\mathcal{L}_{\mathrm{BCE}} = -\frac{1}{|\mathcal{Q}|} \sum_{\mathbf{x} \in \mathcal{Q}} \Big[ & \, y(\mathbf{x}) \log \hat{v}(\mathbf{x}) + \big(1-y(\mathbf{x})\big) \log \big(1-\hat{v}(\mathbf{x})\big) \Big],
\end{aligned}
\end{equation}
where $y(\mathbf{x})=1$ for $\mathbf{x} \in \mathcal{Q}^+$ and $y(\mathbf{x})=0$ for $\mathbf{x} \in \mathcal{Q}^-$.After training, the predicted visible surface is used to recover the relative transformation by minimizing the chamfer distance.

\subsection{Training Data Synthesis}
\label{subsec:data}

Training requires paired examples of observed partial point clouds and their corresponding mesh-aligned visible surfaces. For each object, we pre-render depth maps from 60 uniformly sampled viewpoints to ensure coverage of diverse observation conditions. At each training iteration, we randomly select 1--3 viewpoints per object and back-project the depth maps into the mesh coordinate frame using the known camera poses. This yields a mesh-aligned partial point cloud $\mathcal{P}_{\text{aligned}}$.

To synthesize the network input, we apply a random similarity transformation to $\mathcal{P}_{\text{aligned}}$:
\begin{equation}
\hat{\mathcal{P}}_{\text{obs}} = \{ s R p + t \mid p \in \mathcal{P}_{\text{aligned}} \},
\end{equation}
where $s \in \mathbb{R}^+$, $R \in SO(3)$, and $t \in \mathbb{R}^3$ are randomly sampled, followed by normalization to a unit sphere. We refer to $\hat{\mathcal{P}}_{\text{obs}}$ as a training proxy for real observations. Since the ground-truth alignment is known by construction, the network can be trained in a self-supervised manner to recover $\mathcal{P}_{\text{aligned}}$ from $\hat{\mathcal{P}}_{\text{obs}}$ conditioned on the mesh.

\begin{figure*}[htbp]
    \centering
    \includegraphics[page=1]{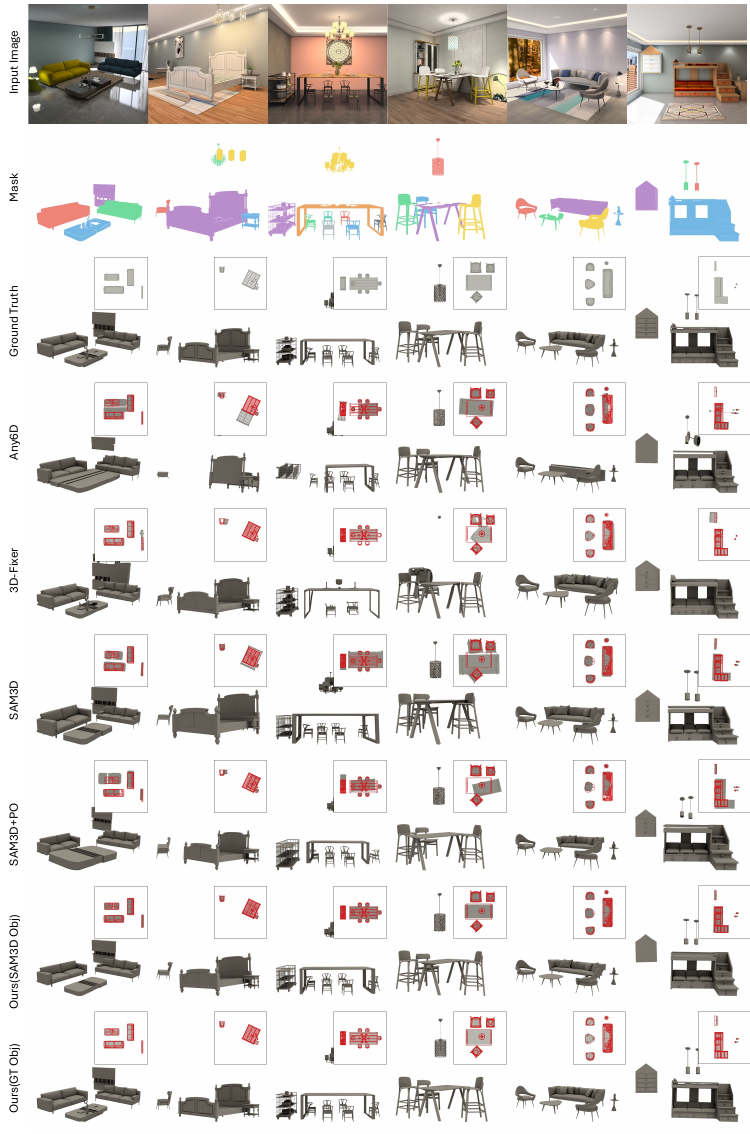}
    \caption{Qualitative comparison of scene-level layout predictions (The rightmost column uses MoGe Depth, while the others use GT-Depth.). Each row shows a scene generated by different methods, with the ground-truth layout shown for reference (red outline). Our method produces layouts that closely align with the ground truth, while baselines exhibit noticeable errors.}
    \label{fig:gt_depth}
\end{figure*}



\section{Experiments}
\label{subsec:setup}

\subsection{Experimental Setup}
\noindent\textbf{Datasets and Benchmarks.}
We evaluate our method on 3D-Future~\citep{3dfuture}, a scene dataset composed of numerous object layouts that can be decomposed into individual meshes. For training, we select 130,000 high-quality meshes from Objaverse~\citep{objaverse}, training with learning rate of 1e-4 for 30000 steps. We evaluate on a randomly sampled subset of 1,000 scenes from 3D-FUTURE test set. Ablation study are conducted on 250 scenes randomly drawn from this subset.

\noindent\textbf{Baselines.}
We compare our method against several paradigms for scene-level layout prediction. \textbf{3DFixer} performs in-place generation conditioned on a reconstructed point cloud of the target object, directly producing a surface conforming to the observed geometry. \textbf{SAM3D} generates objects in canonical space and employs a layout denoising network to predict translation, rotation, and scale in the world coordinate frame. \textbf{SAM3D+PO} (Pose Optimization) initializes from SAM3D predictions and refines the pose by optimizing rendering over the mask and and perform rigid registration between the observed point cloud and the complete point cloud. This refinement aligns the \emph{complete} generated mesh against the \emph{observed partial} point cloud. \textbf{Any6D} is a recent 6D pose estimation framework and we utilize the 3D objects generated by SAM3D as the reference shapes and apply the alignment scheme from Any6D to align the predicted depth and objects for scene-level layout prediction. In contrast, our pose estimation operates directly between two partial point sets without relying on rendering loss and SAM3D initialization: the network-predicted visible surface and the observed partial cloud.

\noindent\textbf{Metrics.}
We jointly assess reconstruction fidelity and layout accuracy using Chamfer Distance (CD) and F-score at $\tau = 0.05$ and $\tau = 0.10$, computed over concatenated object-surface point clouds ($N = 4096$ per object via triangle-area-weighted sampling). To remove global unit scale while preserving relative placement, we apply a single isotropic normalization $x' = (x - c_{\mathrm{GT}}) \cdot 1.9 / s_{\mathrm{GT}}$ derived from the ground-truth axis-aligned bounding box to both clouds, without any subsequent registration (no ICP, rotation fitting, or per-object alignment). Translation, rotation, scale, and object-layout errors therefore remain fully observable in the reported metrics. For ease of reading, the metrics Chamfer Distance (CD) and F-score mentioned below are reported as values scaled by 100.


\begin{table*}[htbp]
    \centering
    \caption{Scene-level quantitative comparison under different depth inputs. All values are multiplied by 100. Best in bold, second best underlined.}
    \label{tab:scene_comparison}
    \resizebox{\textwidth}{!}{%
    \begin{tabular}{l l c c c c c c}
        \toprule
        Depth & Metric & Any6D & 3DFixer & SAM3D & SAM3D+PO & Ours (SAM3D obj.) & Ours (GT obj.) \\
        \midrule
        \multirow{3}{*}{GT}
            & CD $\downarrow$ & 1.04 & 23.35 & 0.98 & 1.51 & \underline{0.36} & \textbf{0.23} \\
            & F@0.05 $\uparrow$ & 85.18 & 85.65 & 75.01 & 74.26 & \underline{90.38} & \textbf{95.60} \\
            & F@0.10 $\uparrow$ & 92.81 & 89.63 & 90.73 & 87.05 & \underline{97.01} & \textbf{98.27} \\
        \midrule
        \multirow{3}{*}{MoGe}
            & CD $\downarrow$ & 2.87 & 23.43 & 1.31 & 1.92 & \underline{0.77} & \textbf{0.70} \\
            & F@0.05 $\uparrow$ & 74.14 & 76.51 & 71.40 & 65.79 & \underline{78.53} & \textbf{81.94} \\
            & F@0.10 $\uparrow$ & 86.80 & 86.48 & 88.77 & 83.48 & \underline{92.40} & \textbf{93.38} \\
        \bottomrule
    \end{tabular}%
    }
\end{table*}

\subsection{Comparison}
\label{subsec:comparison}

Table~\ref{tab:scene_comparison} reports scene-level results under ground-truth depth and MoGe-predicted depth. Across all metrics and settings, our method consistently outperforms all baselines, as illustrated in Figure~\ref{fig:gt_depth}. More results can be found in supplementary material.

\noindent\textbf{Ground-truth depth.}
We use ground-truth geometry as input, which rules out the effect of imperfect geometry estimation, allowing us to observe and compare the output quality of different methods. 3D-Fixer struggles to consistently produce high-quality geometry, and may even fail to generate under conditions such as complex occlusion (i.e., no complete mesh can be extracted), with a CD of 23.35---over an order of magnitude worse than ours. SAM3D generates reasonably but has limited layout accuracy, with F@0.05 of 75.01 versus our 90.38. SAM3D+PO degrades performance versus SAM3D alone. We attribute this to two factors. First, rendering-based alignment is inherently scale-ambiguous: a larger, farther object and a smaller, nearer object can project to nearly identical rendered images, so rendered loss cannot constrain the true scale. Second, its geometric loss aligns a complete mesh to a partial observation; mesh regions absent from the observed point cloud introduce a systematic bias, preventing convergence to correct surface correspondences. Any6D, aligning SAM3D-generated objects with observed depth via its pose and size estimation, achieves a competitive F@0.05 of 85.18, outperforming SAM3D. However, its CD remains high at 1.04, indicating its mechanism still struggles to recover fine-grained geometric details.

In stark contrast, our method achieves exceptional reconstruction fidelity and layout accuracy, driving the CD down to an impressive 0.36 and pushing the F@0.10 to 97.01. This remarkable performance stems from the core strength of our conditional visible-surface reconstruction paradigm. By directly predicting the mesh-aligned visible surface, our network establishes highly precise, dense partial-to-partial correspondences with the observed point cloud. Consequently, under the perfect geometric guidance of GT depth, our method perfectly anchors the generated objects into the scene, demonstrating that our partial-to-partial alignment mechanism is theoretically sound and practically optimal for leveraging high-quality depth observations.

\noindent\textbf{MoGe-predicted depth.}
Under estimated depth input, the performance gap widens further. 3DFixer remains severely degraded, while Any6D suffers from a significant performance drop, with its CD soaring to 2.87 and F@0.05 dropping to 74.14. This highlights that the alignment scheme in Any6D is highly sensitive to depth noise, likely due to scale ambiguity and the lack of robust partial-to-partial matching under noisy observations. SAM3D+PO again fails to improve upon its initialization, consistent with the inherent limitations of full-to-partial alignment under noisy depth. With SAM3D-generated objects, our method achieves a CD of 0.77, a 41\% relative improvement over SAM3D and a 60\% improvement over SAM3D+PO. Notably, our training does not incorporate MoGe-predicted depth for data augmentation, unlike 3DFixer and SAM3D. The robust performance under noisy depth therefore stems from the inherent resilience of our conditional visible-surface reconstruction paradigm rather than from domain-specific augmentation.


\noindent\textbf{Analysis with upstream quality.} 
To decouple our core alignment module from upstream generation artifacts, we evaluate our method under a perfect setting (i.e., using GT depth and GT objects). This configuration establishes the representational upper bound of our framework, purely reflecting the geometric representation capacity and optimization efficacy of our conditional visible-surface reconstruction. Under these idealized geometric conditions, our method achieves exceptional precision, with CD dropping to 0.23 and F@0.10 soaring to 98.27\%. These near-perfect metrics validate the efficacy of our visibility-aware supervision, demonstrating that our network possesses the capacity to establish highly accurate partial-to-partial correspondences when freed from upstream noise. Our method already attains state-of-the-art placement accuracy with current upstream outputs, while the consistent gains observed when using higher-fidelity geometry demonstrate that our framework can readily benefit from further advances in 3D generation and reconstruction.

\begin{table}[htbp]
    \centering
    \caption{Ablation study. All values are multiplied by 100.}
    \label{tab:ablation}
    \resizebox{\columnwidth}{!}{%
    \begin{tabular}{@{}llccccc@{}}
        \toprule
        Depth & Metric & Shared-Encoder & Attn-Swap (w/o Aug.) & Ours (w/o Vis. RCS) & Ours (full) \\
        \midrule
        \multirow{2}{*}{GT} 
            & CD $\downarrow$ & 0.57 & 1.90 & 0.52 & \textbf{0.34} \\
            & F@0.10 $\uparrow$ & 93.86 & 81.75 & 96.65 & \textbf{96.69} \\
        \midrule
        \multirow{2}{*}{MoGe} 
            & CD $\downarrow$ & 0.99 & 2.07 & 0.87 & \textbf{0.74} \\
            & F@0.10 $\uparrow$ & 88.53 & 76.75 & 91.84 & \textbf{92.14} \\
        \bottomrule
    \end{tabular}%
    }
\end{table}

\subsection{Ablation Study}
\label{subsec:ablation}

We conduct ablation experiments to isolate the contribution of each key design choice. Results under both depth settings are reported in Table~\ref{tab:ablation}.


\noindent\textbf{The necessity of separating encoding of mesh and observation.}
We compare against a \textbf{Shared-Encoder} variant that uses one trainable encoder for both the complete mesh and the observed partial cloud, instead of a frozen mesh encoder and a separate observation encoder. As shown in Table~\ref{tab:ablation}, it degrades under both settings: GT CD 0.34$\to$0.57 and F@0.10 96.69$\to$93.86; MoGe CD 0.74$\to$0.99 and F@0.10 92.14$\to$88.53. This confirms that separate encoding is essential to prevent the partial, noisy observation from disturbing the well-learned mesh prior.

\noindent\textbf{Attention Swap.}
Reversing the attention direction (\textbf{Attn-Swap}), where the observation latents serve as queries and the mesh latents provide context, results in a dramatic performance drop: CD increases from 0.349 to 1.905 under GT depth and from 0.787 to 2.071 under MoGe depth. This confirms that retaining the mesh as the primary representation---with the observation acting as a modulator---is essential for preserving the strong geometric prior learned by the mesh decoder.

\noindent\textbf{The necessity of Radius-Constrained Sampling.}
Removing the exclusion radius (\textbf{Ours (w/o Vis.\ RCS)}) degrades performance, with CD increasing from 0.349 to 0.520 under GT depth and from 0.787 to 0.869 under MoGe depth. This occurs because negative samples crowd near the visible/invisible boundary without a radius constraint, creating ambiguous supervision. Introducing a moderate exclusion radius establishes a clear buffer zone that explicitly separates visible and invisible regions, providing sharp supervision that stabilizes training and improves the fidelity of the predicted visible surface.



\section{Conclusion}
\label{sec:conclusion}
We introduce ScenePlacer, a scene-level object pose estimation method that formulates the observation-to-generation correspondence by learning a visibility field. Given a generated mesh and an observed partial point cloud, ScenePlacer predicts a visibility field defined in object space, extracts the visible surface on the generated mesh, and recovers the pose through partial-to-generation alignment. The method leverages a pretrained 3D shape representation, 3DShape2VecSet, uses visibility-aware supervision, and can be trained solely on object-level data without scene-level pose annotations. Experiments on 3D-FUTURE demonstrate consistent improvements in geometric fidelity and layout accuracy over existing baselines under both ground-truth and estimated depth, while the framework remains compatible with future advances in 3D generation and monocular geometry estimation.

\bibliography{iclr2027_conference}
\bibliographystyle{iclr2027_conference}

\appendix
\clearpage
\setcounter{page}{1}
\section{Appendix}

\begin{figure}[htbp]
    \centering
    \includegraphics[page=1, width=\textwidth]{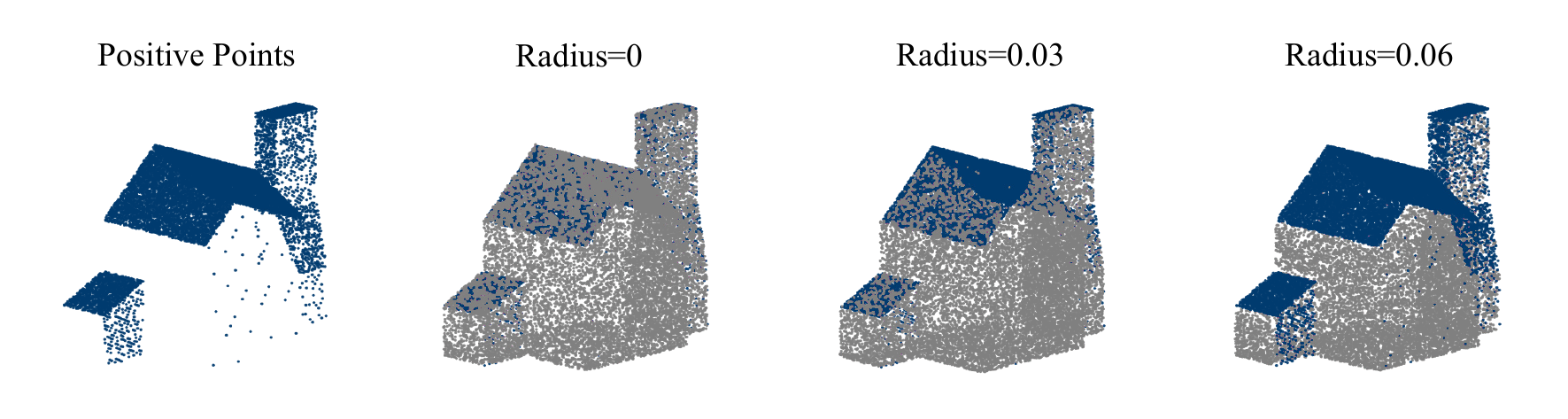}
    \caption{Visualization of positive (visible, blue) and negative (invisible, gray) point sampling under different radius.}
    \label{fig:vis_margin}
\end{figure}

\begin{figure}[htbp]
    \centering
    \includegraphics[page=1, width=\textwidth]{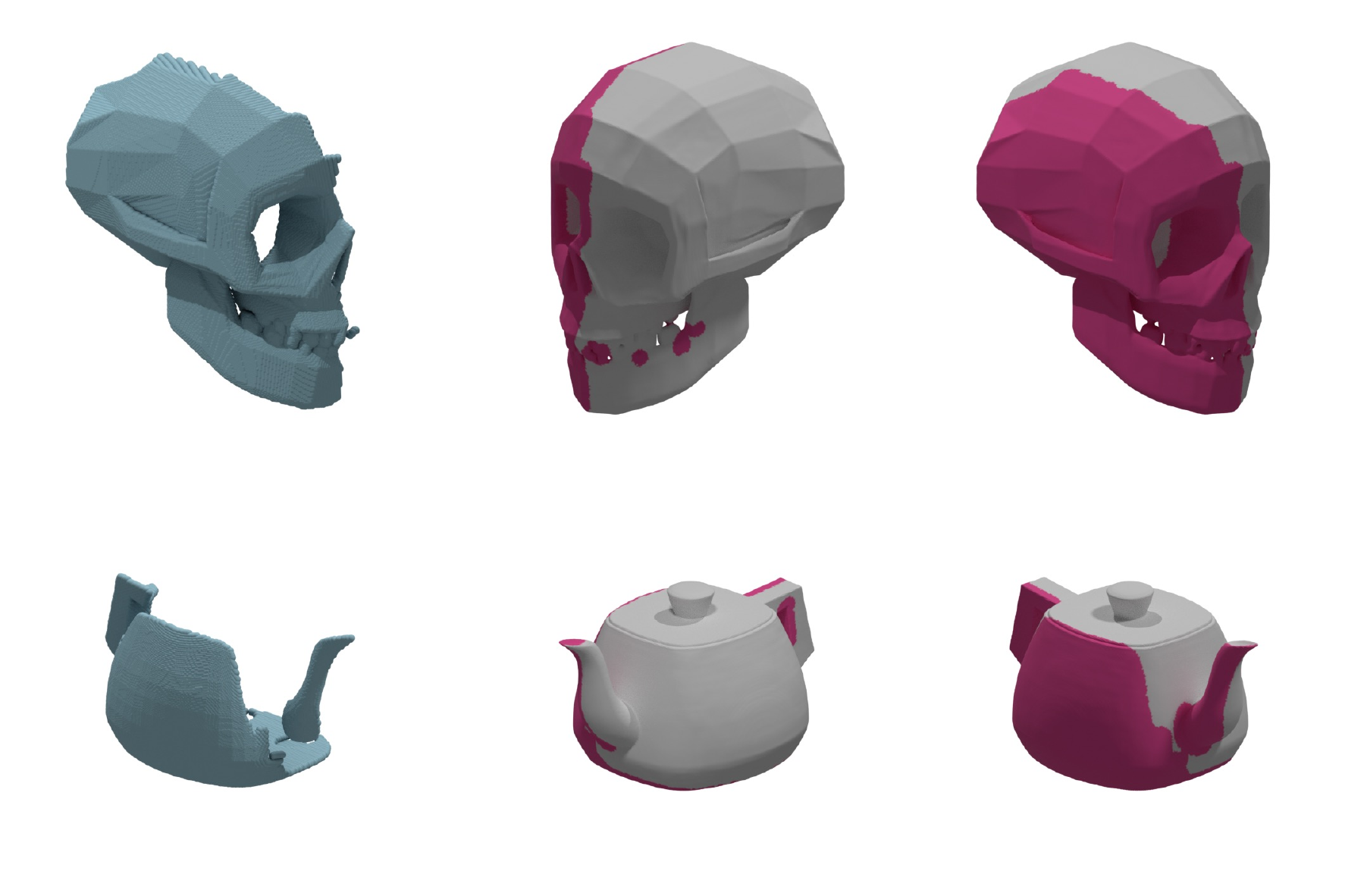}
    \caption{Visibility field visualization.}
    \label{fig:vis_field}
\end{figure}

\begin{figure*}[htbp]
    \centering
    \includegraphics[page=1, width=\textwidth]{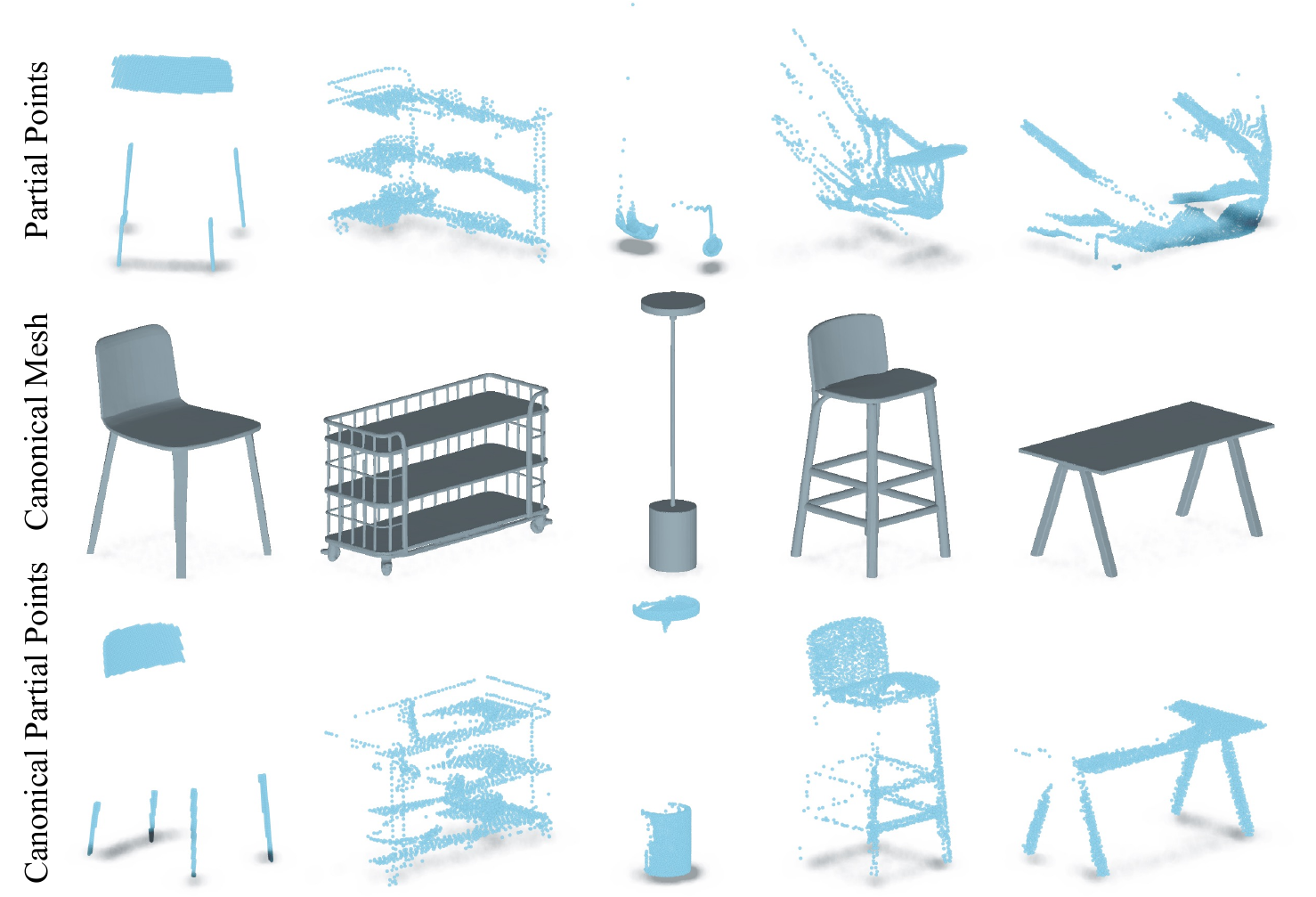}
    \caption{Visibility positive prediction. Given an observed partial point cloud and a generated mesh, the network predicts per-point visibility on the mesh surface and we visualize all positive points in blue.}
    \label{fig:inputoutput_vis}
\end{figure*}

\subsection{The Use of Large Language Models (LLMS)}
We use Large Language Models (LLMs) strictly for minor language editing—grammar and readability, not for method design or experiments. All technical contributions, including the methodology, equations, and results, are solely the work of the authors.

\subsection{More Implementation Details}
We use the pretrained Hunyuan3D-2.1 ShapeVAE~\citep{hunyuan3d} as a point-cloud shape encoder. The original encoder is configured with a 1024-dimensional transformer width, 16 attention heads, 8 encoder layers, Fourier positional features with 8 frequencies, and a 64-dimensional latent space. The pretrained model uses 4096 latent queries and an input surface size of 81,920 points; for efficiency, we reduce the number of latent queries to 512 while retaining the full input surface, resulting in an effective downsampling ratio of 160. The full-shape encoder is frozen and used to provide canonical shape tokens. A second encoder with the same initialization is used for partial observations. In the main experiments, the pre-KL encoder tokens are used as partial-shape context. The resulting Hunyuan tokens are injected into the occupancy decoder through partial-condition cross-attention.

\subsection{More Ablation Study}

\noindent\textbf{Radius Constraint Discussion.}
Incorporating a radius constraint mechanism has significantly improved the sampling of invisible points used for supervision. We visualized the distribution of sampled invisible points based on different radius settings, as shown in Figure~\ref{fig:vis_margin}: if the radius is too small, the invisible points intermingle with visible points—making it difficult to form a continuous surface and causing the field to fragment under supervision—whereas an excessively large radius results in a lack of supervision at the edges. We further visualized the constructed visibility field on the mesh, where the pink areas represent visibility and the white areas represent invisibility; it can be seen that our visibility field effectively captures the relationship between visible and invisible regions formed on the surface through depth back-projection, as illustrated in Figure~\ref{fig:vis_field}.


\begin{table}[htbp]
    \centering
    \caption{Ablation study. All values are multiplied by 100.}
    \label{tab:supp_ablation}
    \resizebox{0.5\columnwidth}{!}{%
    \begin{tabular}{@{}llcc@{}}
        \toprule
        Depth & Metric & Only Pose Solver & Ours \\
        \midrule
        \multirow{2}{*}{GT} 
            & CD $\downarrow$ & 1.32 & \textbf{0.34} \\
            & F@0.10 $\uparrow$ & 86.02 & \textbf{96.69} \\
        \midrule
        \multirow{2}{*}{MoGe} 
            & CD $\downarrow$ & 1.71 & \textbf{0.74} \\
            & F@0.10 $\uparrow$ & 81.34& \textbf{92.14} \\
        \bottomrule
    \end{tabular}%
    }
\end{table}

\noindent\textbf{Only Pose Solver.}
We have already conducted a similar experiment by comparing our method with SAM3D with pose optimization. The pose optimization first preprocesses the generated mesh and checks mask reliability, then performs a heuristic alignment using scale and translation estimated from the point map. ICP then refines the alignment, and finally a render-and-compare stage optimizes the pose via silhouette matching.

To rule out other influencing factors, we conduct an ablation study that uses only a pose solver to align the mesh with the observed partial point cloud. As shown in Table~\ref{tab:supp_ablation}, direct alignment without visibility field prediction leads to severe degradation, further demonstrating the effectiveness of our method.

\subsection{Limitations and Future Work}
\label{subsec:limitations}

Despite our placement method's strong performance, its accuracy is bounded by the quality of observed partial points and generated object meshes. Feed-forward RGB depth estimation still trails traditional geometry reconstruction, and image-conditioned object generation, though silhouette-consistent, lacks robustness to multi-view inputs. These are fundamental challenges for the 3D community, and improving underlying geometry foundation models is essential. Future work will focus on enhancing generation methods' robustness to diverse inputs and improving the geometric priors our placement module builds upon.

\subsection{More Visualization Results}

\begin{figure}[htbp]
    \centering
    \includegraphics[page=1, width=\textwidth]{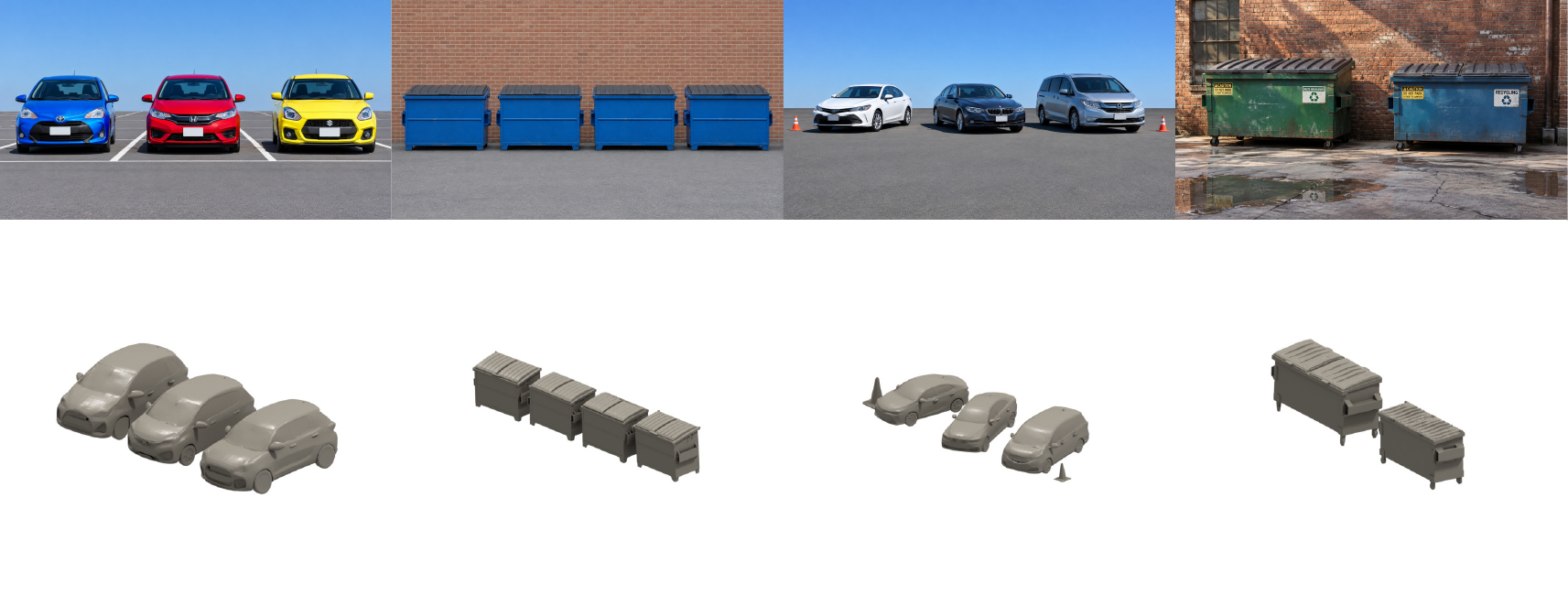}
    \caption{Generalize to outdoor scenes.}
    \label{fig:outdoor}
\end{figure}

\begin{figure}[htbp]
    \centering
    \includegraphics[page=1, width=\textwidth]{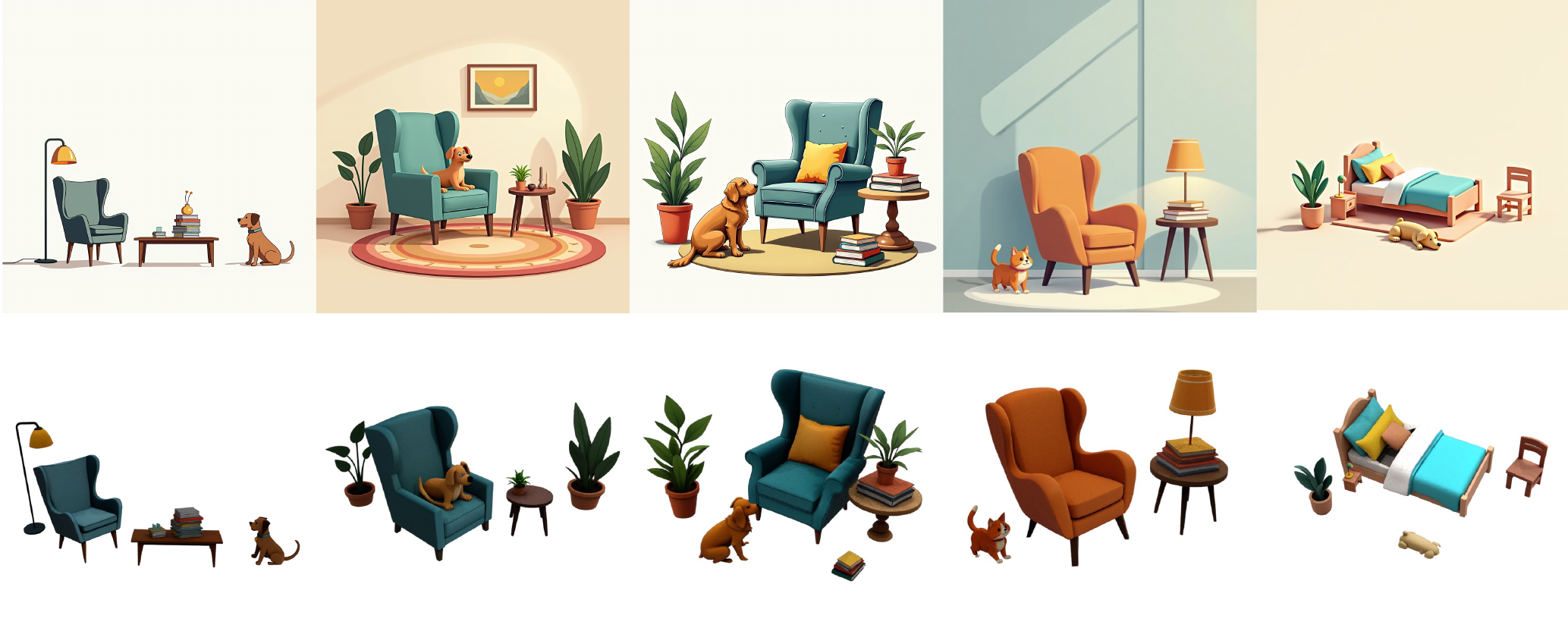}
    \caption{Generalize to cartoon style.}
    \label{fig:cartoon}
\end{figure}

We first visualize some cases of the input and output of our method in Figure~\ref{fig:inputoutput_vis}. Since our method is trained on object-level data, it is insensitive to scene content; it not only applies to indoor scenes but also generalizes to outdoor scenes (Figure~\ref{fig:outdoor}), and even extends to cartoon-style environments (Figure~\ref{fig:cartoon}).

\begin{figure}[htbp]
    \centering
    \includegraphics[page=1, width=\textwidth]{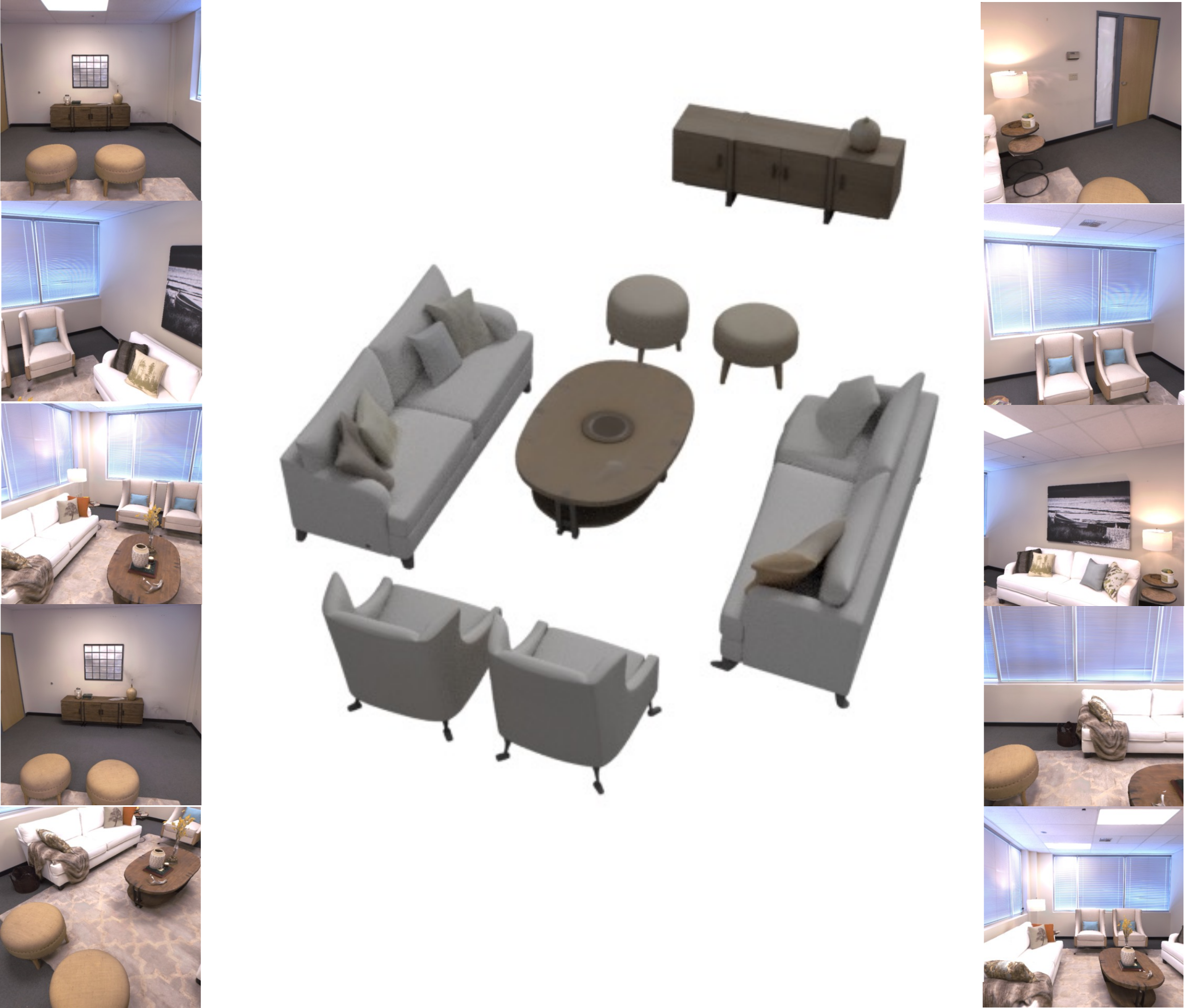}
    \caption{Generalize to multi-view input.}
    \label{fig:replica}
\end{figure}

\begin{figure}[htbp]
    \centering
    \includegraphics[page=1, width=\textwidth]{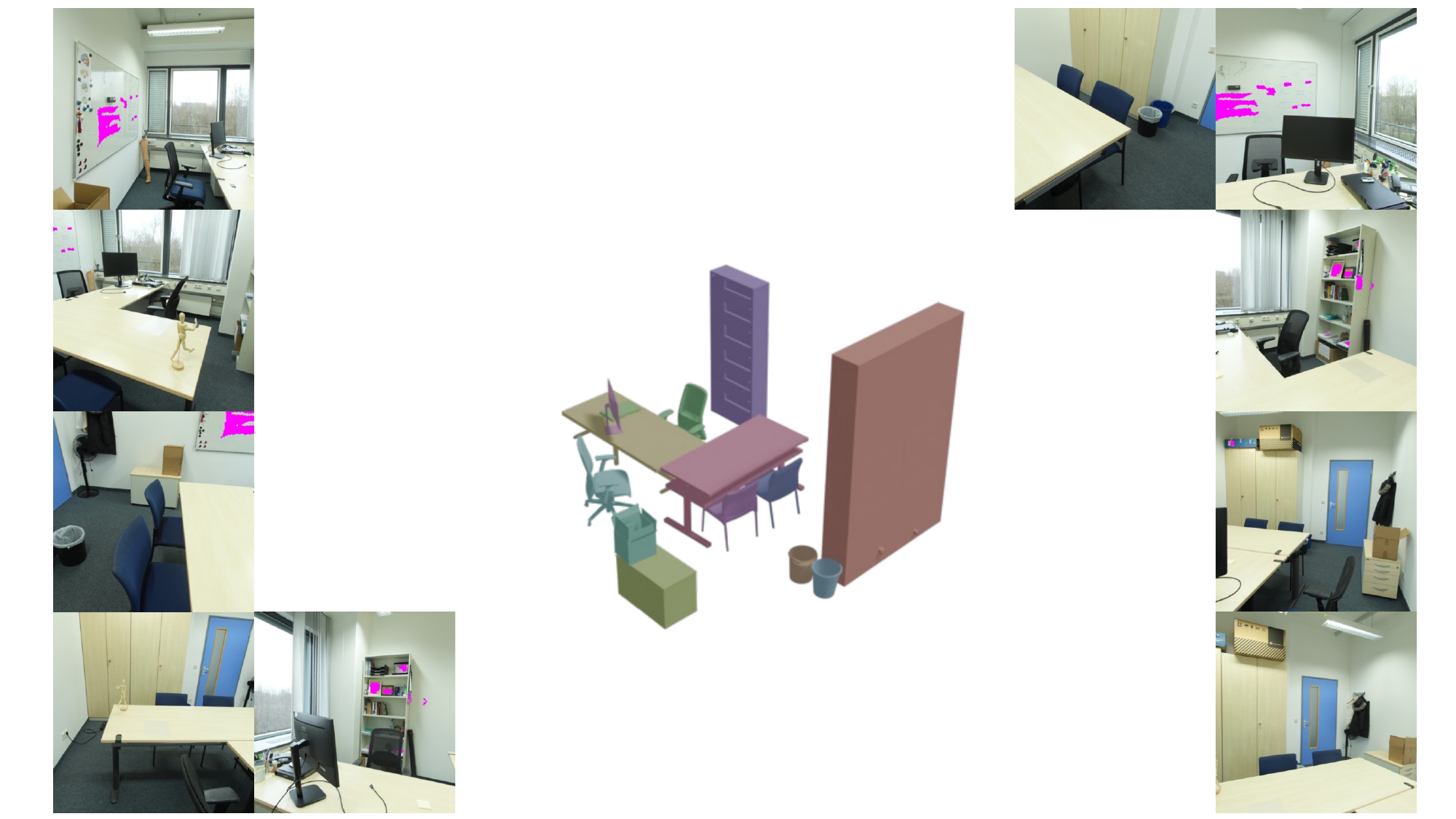}
    \caption{Generalize to multi-view input.}
    \label{fig:scannetpp1}
\end{figure}

\begin{figure}[htbp]
    \centering
    \includegraphics[page=1, width=0.9\textwidth]{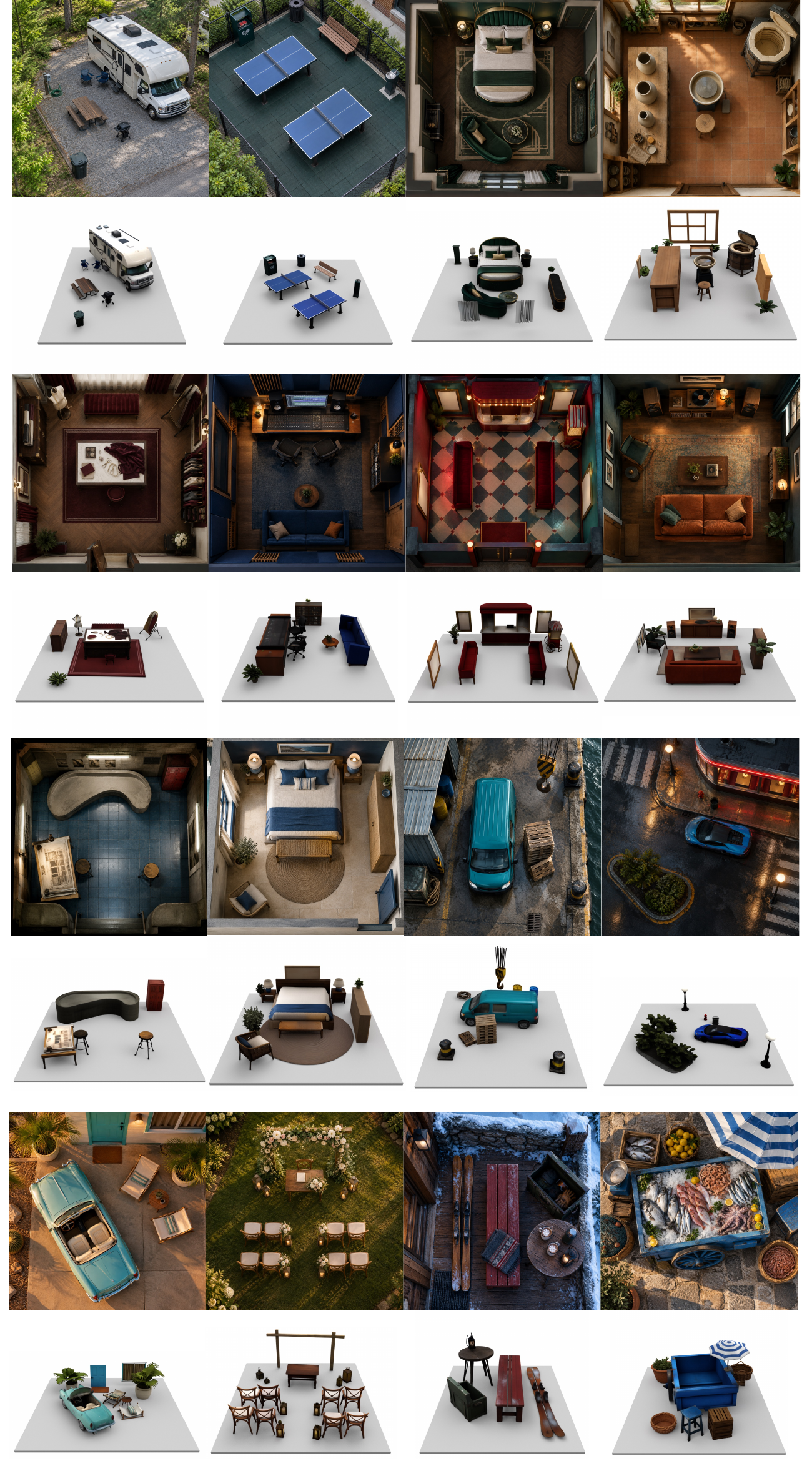}
    \caption{More visualization results.}
    \label{fig:moreresults}
\end{figure}

Moreover, our method can generalize to multi-view inputs. We estimate the geometry via VGGT-Omega~\citep{vggtomga} and generate the object via MV-SAM3D~\citep{mvsam3d}. However, existing generative models struggle to directly handle multi-view inputs, so we rely on successfully generated cases to demonstrate the generalization ability of our method under multi-view inputs, as shown in Figure~\ref{fig:replica} and Figure~\ref{fig:scannetpp1}. We further tested more complex scenarios; since our method places each object individually, it is not sensitive to scene complexity, and we continue to demonstrate excellent generalization capabilities in more complex indoor and outdoor environments and more visualization results can be found in Figure~\ref{fig:moreresults}. We encourage readers to view our accompanying video, which shows the rotation of scenes rendered.

\end{document}